\documentclass[runningheads]{llncs}
\usepackage{graphicx}
\usepackage{amsmath,amssymb} 
\usepackage{color}
\usepackage{multirow}
\usepackage{amsmath}
\usepackage{amsfonts}
\usepackage{algorithm}
\usepackage{algorithmic}
\usepackage{multirow}
\usepackage{colortbl}
\usepackage{color}
\usepackage[table]{xcolor}
\usepackage{epigraph}
\usepackage{graphicx}
\usepackage{caption}
\usepackage{subfigure}
\usepackage{hyperref}
\usepackage{float}
\usepackage{longtable}
\usepackage{pdfpages}
\usepackage{pdflscape}
\usepackage[acronym,toc]{glossaries}
\usepackage{setspace}
\usepackage{wrapfig}
\usepackage{amssymb}
\usepackage{stmaryrd}
\usepackage{array}
\usepackage{enumerate}
\usepackage{paralist}
\usepackage{tabularx}
\usepackage{epsfig}
\usepackage{array}

\def\etc{\emph{etc. }} \def\vs{\emph{vs. }}
 
\def\etal{\emph{et~al.}}

\begin{document}
\pagestyle{headings}
\mainmatter

\def\ACCV20SubNumber{***}  

\title{Depth Completion using Piecewise Planar Model} 
\titlerunning{Piecewise Planer Depth}
%

%
\authorrunning{Zhong et al.}
%
\author{Yiran Zhong$^{1}$, Yuchao Dai$^{2}$, Hongdong Li$^{1}$}
\institute{$^{1}$Australian National University, 
$^{2}$Northwestern Polytechnical University\\
\tt\small{\{yiran.zhong, hongdong.li\}@anu.edu.au}, daiyuchao@nwpu.edu.cn}

\graphicspath{{Figures/}}
\maketitle

\begin{abstract}
A depth map can be represented by a set of learned bases and can be efficiently solved in a closed form solution~\cite{zhong2020efficient}. However, one issue with this method is that it may create artifacts when colour boundaries are inconsistent with depth boundaries. In fact, this is very common in a natural image. To address this issue, we enforce a more strict model in depth recovery: a piece-wise plannar model. More specifically, we represent the desired depth map as a collection of 3D planar and the reconstruction problem is formulated as the optimization of planar parameters. Such a problem can be formulated as a continuous CRF optimization problem and can be solved through particle based method (MP-PBP) \cite{Yamaguchi14}. Extensive experimental evaluations on the KITTI visual odometry dataset show that our proposed methods own high resistance to false object boundaries and can generate useful and visually pleasant 3D point clouds.
\end{abstract}

\section{Introduction}
Autonomous driving requires the vehicles to efficiently sense and understand the surrounding 3D world in real time. Currently, most of the autonomous vehicles (Google, Uber, Ford, Baidu, etc) are equipped with high-end 3D scanning systems such as Velodyne which could provide accurate 3D measurements that are critical to the autonomous vehicles' decision making and planning. However these high-end 3D scanning systems such as Velodyne LIDAR sensors are quite expensive with the cost comparable to the cost of the whole vehicle, which may hinder their admittance to the global consumer market.

To effectively reduce the cost in surrounding 3D world sensing, a natural and cost efficient alternative would be using passive sensors such as monocular cameras or stereo cameras, which could not only provide 3D measurements (by using structure-from-motion (SfM) or simultaneous localization and mapping (SLAM)) but also semantic information (not available from the point clouds). However, the computer vision based system is either not robust (different weather conditions could result in dramatically different vision measurements) or inaccurate (compared with LiDAR based 3D scanning). 

\begin{figure}[!htp]
 \centering
    \includegraphics[width=1\columnwidth]{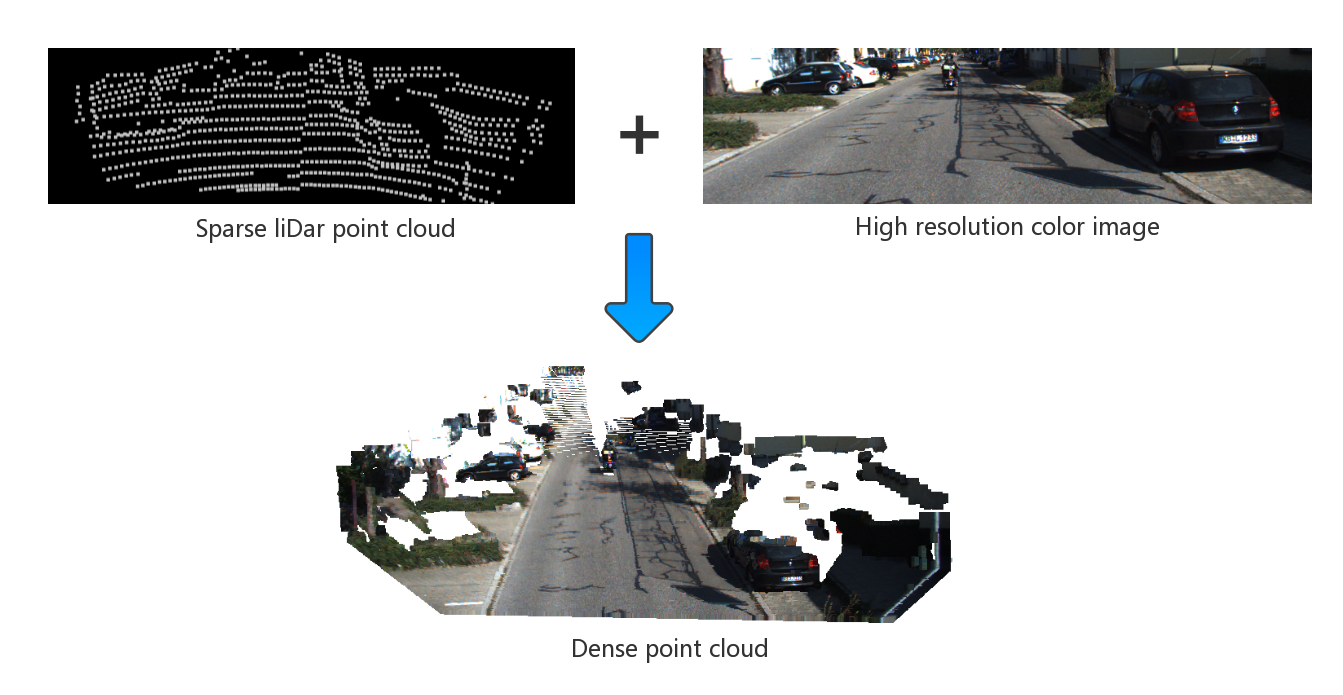} 
  \caption[Problem definition.]{\label{fig:sparse-dense-concept} Conceptual illustration of the problem of predicting dense 3D point clouds/depth maps from sparse LiDar input, where the input is a collection of sparse 3D point clouds and desired output is a dense 3D point cloud/depth map.}
  \label{fig:conceptual-illustration}
\end{figure}

In this paper, we propose to investigate effective fusion to integrate sparse 3D point clouds (low cost) and high-resolution colour images, such that generating dense 3D point clouds comparable to high-end 3D scanning systems. Specifically, the objective of this paper is to ``\textbf{generate/predict a dense depth map and the corresponding 3D point clouds from very sparse depth measurements with/without colour images and colour-depth image datasets}''.  Under our framework, we use a low-resolution LIDAR and a high-resolution colour image to generate dense depth maps/3D point clouds (Fig.~\ref{fig:conceptual-illustration}). This can be understood as using the high-resolution colour image to augment the sparse LIDAR map and predict a dense depth map/3D point clouds. Our proposed methods can not only provide accurate dense depth maps but also provide visually pleasant 3D point clouds, which are critical for autonomous driving in urban scenes. Specifically, we have presented a novel way in perceiving 3D surrounding environments, which owns low-cost compared with the high-end LIDAR sensors and high precision and efficiency compared with the colour camera only solutions. The proposed framework owns great potentials in developing compact and high-resolution LIDAR sensors, at very low cost, for domain-specific applications (e.g. ADAS, autonomous driving).

Our dense depth map prediction framework differs from general depth interpolation methods and depth super-resolution methods in the following aspects: 1) Irregular depth points pattern; 2) Very sparse depth measurements; and 3) Missing real obstacles or generating false obstacles in a certain range is unacceptable. The irregular depth points pattern creates a barrier in applying deep convolutional neural network (CNN) based methods as it requests irregular convolution pattern, which is still an unsolved problem in deep learning. On the other hand, even though many traditional methods \cite{Li2012,ferstl2013b,YangAR2014,Barron2016,Li2016} could deal with irregular patterns, they suffer from the dependency on the strong correlation between colour images and depth maps, where boundaries that only exist in the colour images will mislead the methods and generate false obstacles on road. For those general smooth interpolation methods such as nearest neighbor, bilinear interpolation, bicubic interpolation and \cite{Garcia20101167}, since they neglect the colour images in problem formation, they tend to generate over-smoothed results and miss real obstacles.

In this paper, targeting at handling the above difficulties with existing methods, we propose a dense depth prediction approach that only uses the boundaries in colour image (i.e., the superpixel over-segmentation). We resort to the piecewise planar model for general scene reconstruction, where the desired depth map/point clouds are represented as a collection of 3D planar and the reconstruction problem is formulated as the optimization of planar parameters. The resultant optimization involves the unary term evaluated at the sparse depth measurements, the smoothness term across neighboring planes. Thirdly, as the urban driving scenarios are well structured, we propose a specifically designed model for urban driving scenario called ``cardboard world'' model, i.e., front-parallel orthogonal piecewise planar model, where each segment can only be assigned to either the road plane or a front-parallel object plane orthogonal to the road plane. We formulate the problem as a continuous CRF optimization problem and solve it through particle based method (MP-PBP) \cite{Yamaguchi14}. Extensive experiments on the KITTI visual odometry dataset show that the proposed methods owns high resistance to false object boundaries and can generate useful 3D point clouds without missing obstacles.

\section{Related Work}
Given an incomplete depth map, depth completion task is to fill the missing depth values to get a dense depth map. A colour image that is captured with a camera is often used as the guided image. Since most range sensors can only achieve sparse or semi-dense depth maps, completing missing depth values can increase the quality of depth maps for better applicability. This task has drawn increasing attention recently due to the rising interest of autonomous driving, augmented/virtual reality and robotics. Although this task can be partially addressed by traditional image in-painting techniques, extra knowledge such as depth smoothness, normal smoothness, colour guidance and \etc is yet to be utilized for achieving an accurate dense depth map. 

Depth completion includes three sub-tasks: depth inpainting, depth super resolution and depth reconstruction from sparse samples. Anisotropic diffusion \cite{Perona1990} is a popular method for image inpainting, and it has been successfully adapted to the depth inpainting task in \cite{Miao2012}. Energy Minimization based depth inpainting methods \cite{Liu2013icip,Chongyu2015} take characteristics of depth maps as a regularization term in energy minimization and generate more plausible results. Exemplar-based Filling \cite{Criminisi2004} works well in image inpainting but creates extra challenges in depth inpainting due to the lacking of textures in depth maps. Matrix Completion \cite{Lu2014} assumes that a depth map lies in a low-dimensional subspace and can be approximated by a low-rank matrix. The main goal for depth super-resolution task is to increase the spatial resolution of a depth image to match a high resolution colour image. The low-resolution depth maps are completed in this case. By assuming depth discontinuities are often aligned with colour changes in the reference image, this task can be solved through Markov Random Fields (MRF) \cite{DiebelT05,Andreasson07}. Such methods can be easily adapted to depth reconstruction from sparse samples task as they do not require the depth points to be regularly sampled \cite{DolsonBPT10}. In order to intensively use the structural correlation between colour and depth pairs in colour guided depth super-resolution, non-local means (NLM) was introduced as a high-order term in regularization \cite{Park2011}. \cite{YangAR2014} swaps the Gaussian kernel in the standard NLM to a bilateral kernel to enhance the structural correlation in colour-depth pairs and proposed an adaptive colour-guided auto-regressive (AR) model that formulates the depth upsampling task as AR prediction error minimization, which owns a closed-form solution. A few approaches employ sparse signal representations for guided upsampling making use of the Wavelet domain \cite{Hawe2011}, learned dictionaries \cite{Li2012} or co-sparse analysis models \cite{Gong2014}. \cite{Barron2016} and \cite{Li2016} leverage edge-aware image smoothing techniques and formulate it as a weighted least squares problem while \cite{Li2016} also applied coarse-to-fine strategy to deal with the very sparse situation. \cite{ku2018defense} proposed an efficient depth completion algorithm that achieves the state-of-the-art performance on KITTI depth completion benchmark \cite{Uhrig2017THREEDV} that only uses image processing operations.

With the booming development of deep learning technique, deep networks were introduced to geometry learning tasks such as monocular depth estimation~\cite{Zhong2018ECCV}, stereo matching~\cite{zhong2017self,Zhong2018ECCV_rnn,zhong2020nipsstereo,zhong2020displacementinvariant}, optical flow~\cite{Zhong_2019_CVPR,zhong2020nipsflow}, and achieved state-of-the-art performance on most benchmarks. By treating a depth map as a gray-scale image, image super-resolution networks \cite{Dong2014} can be directly applied \cite{Song2016}. Riegler \etal \cite{Riegler2016} proposed a deeper depth super-resolution network that has faster convergence and better performance. A natural barrier for applying deep methods to other depth completion tasks is the irregular depth pattern as standard convolution layers are designed for regular grid inputs. Sparsity Invariant CNNs \cite{Uhrig2017THREEDV} addresses this problem and tried to handle the sparse and irregular inputs by introducing invalid masks in convolution layers. On the other hand, \cite{jaritz2018sparse} claims that standard convolution layers can handle sparse inputs of various densities without any additional mask input. In \cite{semantic2016}, it proves that leveraging semantic information can improve the depth completion performance and vice versa~\cite{zhong20183d}. In the recent stereo-LiDAR fusion work~\cite{Cheng_2019_CVPR}, they also apply the piecewise planner model as a soft constraint in their network loss functions.

\section{Approach}
Real world driving scenarios generally consist of road, surrounding buildings, vehicles, pedestrians and etc., which can be well approximated with piece-wise planar model in 3D representation. By representing the traffic scenes with piece-wise planar model, we are able to exploit the structural information in the scenes and the number of variables can be greatly reduced (For each plane, we only need 3 parameters to represent.). In this way, the number of measurements of sparse depths could be greatly reduced, which enables us the ability to work with very sparse LIDAR measurements. Furthermore, the parametric model owns the ability to handle outlying LIDAR measurements. 

\subsection{An overview of the proposed method}
\begin{figure}[!htp]
\begin{center} 
 \subfigure{
\includegraphics[width=1\columnwidth]{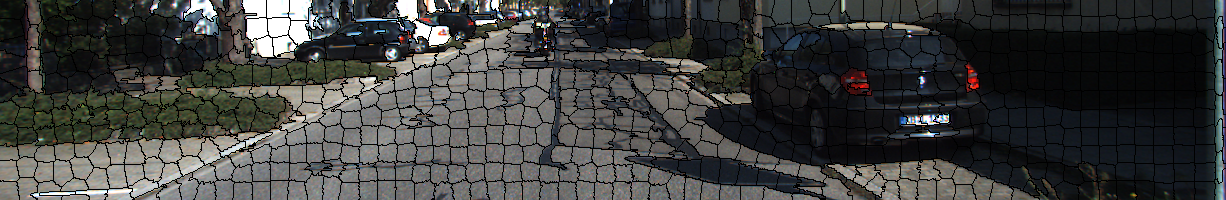} }
 \subfigure{
   \includegraphics[width=1\columnwidth]{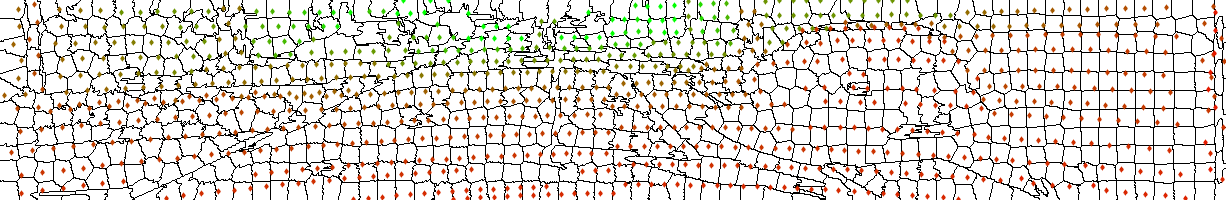} }
 \caption[Inputs of our algorithm.]{\label{fig:input} \textbf{Inputs of our algorithm implementation.} Top: The SLIC over-segmentation of the input image. Bottom: the input of our algorithm: the super-pixel segmentation and sparse depth measurements. Note that as the depth measurements are very sparse, there are only a few depth measurements in each super-pixel and there are considerable super-pixels where no depth measurements are available.}
 \end{center}
\end{figure}

\begin{figure}[!htp]
\centering
\begin{center}
\includegraphics[width=.65\columnwidth]{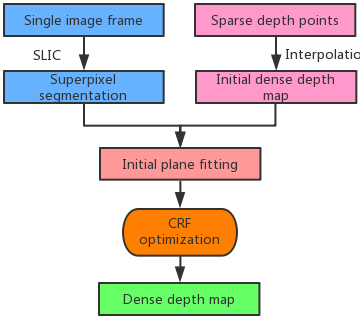}
\caption[Pipeline of our method.]{\label{fig:processcrf}\textbf{Pipeline of our method.} Given high resolution colour image and sparse depth measurements as input, our approach predicts dense depth map with the same resolution as the colour image.}
\end{center}
\end{figure}

A high level description of our method is given as follows: assume the LIDAR sensor and the camera have been geometrically calibrated (both intrinsically and extrinsically).  Given the input of a single frame monocular image with corresponding sparse depth points, we first perform image over-segmentation to obtain fixed number of super-pixels using the SLIC algorithm \cite{SLIC2012}. Then we conduct interpolation on the sparse depth points to generate an initial dense depth map by using the penalized least squares method \cite{Garcia20101167}. This dense depth map is used to provide initial planar parameters for each segment. As shown in Fig.~\ref{fig:input}, on average each superpixel region contains a single depth point measurement. Also note some superpixels do not contain any depth measurement.  After fitting the initial depth measurements inside each superpixel with a plane, we formulate a Conditional Random Field (CRF) to optimize all the plane parameters and recalculate the depth map. A flowchart of our approach is given in Fig.~\ref{fig:processcrf}. 

\subsection{Mathematical Formulation}
Under our piece-wise planar model for the scenes, each segment is well approximated with a plane. In this way, the dense depth prediction problem is transformed to the optimization of the plane parameters. Furthermore, we assume that the boundaries in the depth maps are a subset of the boundaries in the colour images, which enables us the freedom to decide the real depth boundaries from the colour boundaries.


Specifically, let $\mathcal{S}$ denote the set of superpixels and each superpixel $i\in \mathcal{S}$ is associated with a segment $\mathcal{R}_i$ in the image plane and a random variable $\mathbf{s}_i = \mathbf{n}_i$, where $\mathbf{n}_i \in \mathbb{R}^3$ describes the plane parameter in 3D. Our goal is to infer the 3D geometry of each superpixel $\mathbf{s}_i$ given the sparse depth measurements $\mathrm{d}_i(\mathbf{x})$. We define the energy of the system to be the sum of a data term $\varphi_i$ and a smoothness term $\psi_{i,j}$,
\begin{equation}
E(\mathbf{s}) = \sum_{i\in \mathcal{S}}\varphi_i(\mathbf{s}_i) + \sum_{i\sim j}\psi_{i,j}(\mathbf{s}_i,\mathbf{s}_j),
\label{energy}
\end{equation}
where $\varphi$ denotes the data term while $\psi$ denotes the smoothness term. $\mathbf{s} = \{\mathbf{s}_i|i\in \mathcal{S}\}$ and $i\sim j$ denotes the set of adjacent superpixels in $\mathcal{S}$.

\noindent\textbf{Data Term:} The data term encourages the depth measurement to lie on the plane by penalizing the discrepancy between the measurement and the prediction. To better enforce this constraint, we choose the $\ell_2$ norm to amplify the cost. Therefore, our data term can be written as:
\begin{equation}
\varphi_i(\mathbf{s}_i) = \theta_1 \sum_{\mathbf{x}\in \mathcal{X}}\|\widehat{\mathrm{d}}(\mathbf{s}_i,\mathbf{x}) - \mathrm{d}_i(\mathbf{x})\|_2^2,
\label{dataenergy}
\end{equation}
where $\mathcal{X}$ is the set of pixels that has depth measurements. $\widehat{\mathrm{d}}(\mathbf{s}_i,\mathbf{x}) = \frac{-\widetilde{\mathrm{d}}_i}{\mathbf{n}_i^T \mathbf{K}^{-1}\mathbf{x}}$ represents the estimated depth value on a pixel $x$, $\widetilde{\mathrm{d}}_i$ represents the distance between the plane and the origin, $\mathbf{n}_i^T$ is the normal vector of the plane, $\mathbf{K}$ is the camera intrinsic matrix and $\mathbf{x} = (u,v,1)^T $ is the homogeneous representation of the pixel $x$. $\mathrm{d}_i(x)$ is the depth measurement on pixel $x$. 

\noindent
\textbf{Smoothness Term:}
The smoothness term encourages coherence of adjacent superpixels in terms of both depth and orientation. The depth coherence is defined by the difference between the depths of neighboring superpixels' boundaries while the orientation coherence is defined as the difference between the surface normal of neighbouring superpixels. Considering the discontinuity in neighboring superpixels due to scene structure, we use the truncated $\ell_1$ norm to allow discontinuity in both depth and orientation. 


Following \cite{Menze2015CVPR}, our smoothness potential $\psi_{i,j}(\mathbf{s}_i,\mathbf{s}_j)$ can be decomposes as:
\begin{equation}
\psi_{i,j}(\mathbf{s}_i,\mathbf{s}_j) = \theta_2 \psi_{i,j}^{depth}(\mathbf{n}_i,\mathbf{n}_j) + \theta_3 \psi_{i,j}^{orient}(\mathbf{n}_i,\mathbf{n}_j),
\label{smenergy}
\end{equation}
with weights $\theta$ and
\begin{equation*}
\psi_{i,j}^{depth}(\mathbf{n}_i,\mathbf{n}_j) = \sum_{\mathbf{p}\in \mathcal{B}_{i,j}}\rho_{\tau_1}(d(\mathbf{n}_i,\mathbf{p})-\mathrm{d}(\mathbf{n}_j,\mathbf{p})),
\end{equation*}
\begin{equation*}
\psi_{i,j}^{orient}(\mathbf{n}_i,\mathbf{n}_j) = \rho_{\tau_2}(1-|\mathbf{n}_i^T\mathbf{n}_j|/(|\mathbf{n}_i||\mathbf{n}_j|)),
\end{equation*}
where $\mathcal{B}_{i,j}$ is the set of shared boundary pixels between superpixel $i$ and superpixel $j$, and $\rho$ is the robust $\ell_1$ penalty function $\rho_{\tau}(x) = \min(|x|,\tau)$.

\subsection{CRF Optimization}
The optimization of the above continuous CRF defined in Eq.~\eqref{energy} is generally NP-hard. In order to efficiently solve this optimization problem, we discretize the continuous variables and leverage particle convex belief propagation (PCBP) \cite{PCBP2011}, an algorithm that is guaranteed to converge and gradually approach the optimum. It works in the following way: after initialization, for each random variable, particles are sampled around current states. Then these particles act as labels in discretized MRF/CRF that can be solved by any MRF/CRF solving methods such as multi-label graph cut, sequential tree-reweighted message passing (TRW-S) \cite{Kolmogorov2006TRWS} and update the MAP estimation to current solution. The process is repeated for a fixed number of iterations or until convergence.

\begin{algorithm}[t!]
\caption{Solving Eq.~\eqref{energy} via the PCBP}
\label{Algorithm 1}
\begin{algorithmic}
\REQUIRE ~~\\
Superpixels $\mathbf{S}$ and sparse depth measurements $\mathbf{d}$, number of particles $n_p$, number of iterations $n_i$, parameters $\theta_1$, $\theta_2$, $\theta_3$, $\tau_1$, $\tau_2$, $\rho$, $\sigma$. 
 \\ \vspace{0.2cm}
\hspace{-0.3cm}{\bf Initialize:} Initial plane parameters $\mathbf{s}_i$ for each superpixel segment. \\ \vspace{0.2cm}
\WHILE {iteration $< n_i$}
\STATE 1). Sample particles: the first particle for each superpixel is the result of previous iteration, the next $n_p/2$ particles are randomly sampled around the state in the previous iteration and the remaining $n_p/2-1$ particles are sampled from the neighboring superpixels' current states;

\STATE 2). Evaluate the data term \eqref{dataenergy} and the smoothness term \eqref{smenergy};

\STATE 3). Solve the resultant discrete problem with TRW-S and update plane parameters for each superpixel.

\ENDWHILE
\ENSURE ~Plane parameters $\mathbf{S}$, recovered depth map $\mathbf{D}$.
\end{algorithmic}
\end{algorithm}

\textbf{Implementation Details:} A 3D plane is defined by: 
\begin{equation}
aX_i+bY_i+cZ_i+d = 0
\label{eq:abcdplane}
\end{equation}
where $(a,b,c,d)$ are the plane parameters, $(X_i,Y_i,Z_i)$ is the 3D points coordinates that can be computed by
\begin{equation}
X_i = (u_i - C_x)\times Z_i/f,
\end{equation}
\begin{equation}
Y_i = (v_i - C_y)\times Z_i/f,
\end{equation}
where $Z_i = d_i$. $(u_i,v_i)$ is the point coordinate in the image plane, $(C_x,C_y)$ is the camera principal point offset and $f$ is the camera focal length. 

Thus, the $i^{th}$ particle is defined as a $4\times 1$ vector $(a_i,b_i,c_i,d_i)^T$. $a_i,b_i,c_i,d_i$ are independently generated through a normal distribution with standard deviation $\sigma$ and mean $\mu$, where $\sigma$ is given by user setting and $\mu$ is the state in the previous iteration.

Our approach to solve Eq.~\eqref{energy} is outlined in Algorithm 1, where parameters $\theta_1$ $\theta_2$ $\theta_3$ $\tau_1$ $\tau_2$ have been already defined in previous equations, $\rho$ is the decay rate for generating particles and $\sigma$ is a $4\times 1$ vector that contains the standard deviation for the 4 parameters of each particle.

\textbf{Particle Generation:} Instead using of the regular PCBP particle generation scheme, which generates particles only from the MCMC framework, we partially adopt the PMBP \cite{PMBP2014} scheme by adding neighboring plane parameters into the particles, thus the candidate particles are a mixture of MCMC sampling around the previous states and the states from neighboring planes. The advantage of this modification is that it allows neighboring superpixels to fuse together thus decreasing the energy. For example, the road area is often segmented into several segments. By using the regular PCBP particle sampling strategy, each segment's parameters are independently generated by a normal distribution. Even though the smoothness term encourages the planes be close to each other, there will still exist small gap between them as the algorithm could not find exactly the same parameters from its candidate particles. However, in our modified version, the neighboring superpixels can share their parameters and therefore resolve the issue.

In each PCBP iteration, each superpixel has fixed number of particles, and we need to find the sub-optimal combination that has the lowest energy. This problem can be efficiently solved through tree-reweighted max-product message passing (TRW-S) algorithm. The processing time depends on the number of superpixels and the number of particles.

\subsection{Cardboard World Model: A More Constrained Model}
In the above section, we described our piece-wise planar model for solving the dense depth prediction problem, where each plane has three freedoms in 3D space. 

Here we notice that for real-world driving applications, man-made road scenes often have stronger structured information (i.e. stronger prior).  In particular, we realize that modeling a front-view road scene as a combination of ground-plane and many front-parallel obstacles planes will be convenient for the task of drivable free-space detection task. Next, we will show how to infer such a simplified 3D road scene model by using the same method of our CRF--conditional random field framework. 

\begin{figure}[!htp]
\begin{center} 
\includegraphics[width=1\columnwidth]{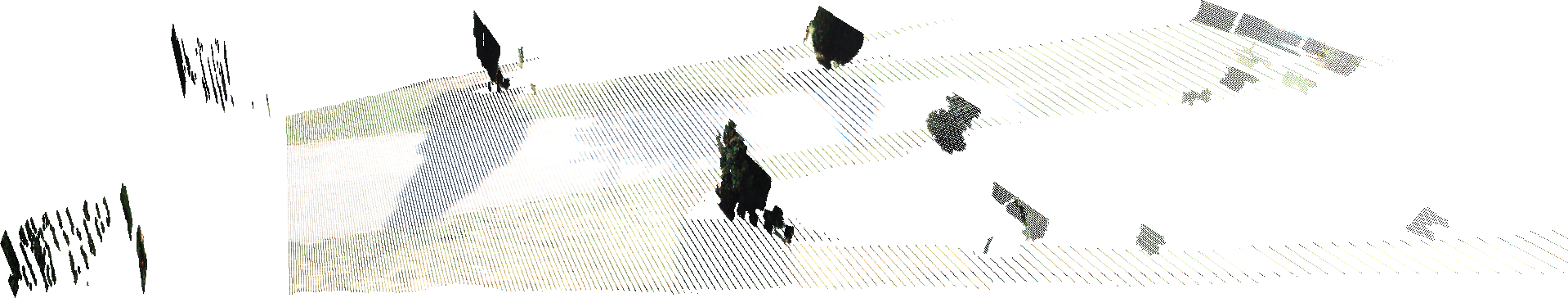} 
 \caption[``Cardboard world'' model.]{\label{fig:cw} \textbf{An example of our ``cardboard world'' model:} Object planes are front parallel planes that orthogonal to the road plane.}
 \end{center}
\end{figure}

Based on these observations, we propose our ``cardboard world'' model for representing the driving scenes. Under the ``cardboard world'' model, there only consist two kinds of planes: the road plane and the object plane. We assume there is only one road plane in the scene and all the other planes are object plane. These two kinds of planes are orthogonal to each other and object planes are front parallel. Fig.~\ref{fig:cw} illustrates an example of our ``cardboard world'' model. 

There are three advantages in replacing the slanted plane model in the piecewise planar model with our proposed ``cardboard world'' model:
\begin{compactenum}[1)]
\item The recovered 3D point clouds are more visually pleasant and the location of each object is more accurate. In some case, when there are two depth points with very different depth values in a single superpixel, the slanted plane model will use a very slanted plane to fit these two points. As a result, the shape of this area will be largely distorted. On the other hand, in the ``cardboard world'' model, it will force the plane to become front parallel therefore maintain the object shape.
\item As a byproduct, this method provides a free-space for autonomous driving vehicles. We embed a binary labeling problem in our task that classifies superpixels into two clusters: road and object. Superpixels labeled as road belong to the free-space.
\item Processing time can be greatly reduced. By applying a front-parallel constraint and orthogonal constraint, the number of parameters to optimize have been greatly reduced thus decreasing the processing time.
\end{compactenum}

\begin{figure}[!htp]
\centering
\begin{center}
\includegraphics[width=.65\columnwidth]{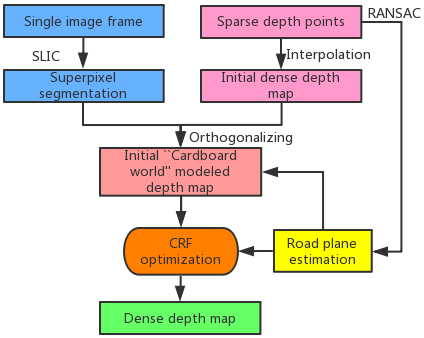}
\caption[Process flow of our method.]{\label{fig:process}\textbf{Process flow of our approach.} Given high resolution colour image and sparse depth measurements as input, our approach predicts dense depth map with the same resolution as the colour image.}
\end{center}
\end{figure}

A diagram of our cardboard model approach is shown in Fig.~\ref{fig:process}.

\textbf{Initialization:} Different from the piecewise planar based method, this new method requires an initial road plane estimation for initialization. To do so, we fit the road plane directly from sparse depth point using RANSAC. After getting the initial dense depth map using \cite{Garcia20101167}, for each superpixel segment, we project all the depth points into 3D and calculate the distance between them and the road plane by summing the Euclidean distance of each point. If the distance is smaller than a given threshold $\epsilon$, we label the superpixel as road plane and fit it with the initial road plane parameters. Otherwise, the superpixel is labeled as object and will be fitted with a front parallel plane with a depth of the mean depth of this superpixel.

The optimization problem remains the same as the previous one. However, since we need to enforce the orthogonal relationship as well as the front parallel constraint, given road plane parameter $(a_r,b_r,c_r,d_r)^T$, the normal of object planes are fixed as well and are equal to $(0, -\frac{c_r}{b_r},1)^T$. Hence there is only one freedom in object planes, i.e., the unknown depth. Note that we naturally embed a binary labeling problem in optimization step since superpixels are divided into two groups: road plane group and object planes group. This labeling problem is jointly solved through PCBP process by adding road plane parameters into particle sets of every superpixels, so that every superpixel has a choice to joint the road plane when the fitting cost is low enough. Our approach to solving Eq.~\eqref{energy} using our ``cardboard world'' model is outlined in Algorithm 2.

\begin{algorithm}[t]
\caption{Solving Eq.~\eqref{energy} under ``cardboard world'' model via the PCBP}
\label{Algorithm 2}
\begin{algorithmic}
\REQUIRE ~~\\
Superpixels $\mathbf{S}$ and sparse depth measurements $\mathbf{d}$, number of particles $n_p$, number of iteration $n_i$, parameters $\theta_1$, $\theta_2$, $\theta_3$, $\tau_1$, $\tau_2$, $\rho$, $\sigma$, $\epsilon$. 
\\ 
\hspace{-0.3cm}{\bf Initialize:} Estimate road plane parameter $\mathbf{s}_d$, initial dense depth map $\mathrm{D}_0$  
\FORALL{$\mathbf{S_i} \in \mathbf{S}$}
    \STATE Project every depth points into 3D;
    \STATE Calculate the sum of Euclidean distance between each 3D point the road plane $\mathbf{s}_d$;    \IF{The sum of Euclidean distance is less than a given threshold $\epsilon$}
      \STATE Fit $\mathbf{S}_i$ with $\mathbf{n}_\mathrm{d}$
    \ELSE
      \STATE Fit $\mathbf{S}_i$ with front parallel plane and orthogonal to $\mathbf{n_{\mathrm{d}}}$
    \ENDIF
\ENDFOR
\WHILE {iteration $< n_i$}
\STATE 1). Sample particles: the first particle for each superpixel is the state in previous iteration, then randomly generate $n_p/2-1$ particles and add $n_p/2-1$ neighboring superpixels' parameters to the particle set and add the last particle with $\mathbf{n_d}$;
\STATE 2). Calculate data energy through \eqref{dataenergy} and smoothness energy through \eqref{smenergy};
\STATE 3). Solve through TRWS and update current MAP estimation ${\bf n}$ to the first particle;\\
\ENDWHILE
\ENSURE ~Plane parameters ${\bf n}$, recovered depth map ${\bf D}$.
\end{algorithmic}
\end{algorithm}

\section{Experiments}
\subsection{Experimental setup}
We perform both quantitative and qualitative evaluations of our methods on the KITTI VO dataset. The KITTI VO dataset consists of 22 sequences 43,596 frames which includes various driving scenarios such as highway, city and country road. It provides stereo images and semi-dense $360^\circ$ LIDAR points. In our experimental setting, we only use the left images and the sparse LIDAR points. Note, we recover the depth for the every 10th frames, thus 4,359 frames are recovered in total. 

We perform evaluation of two subset of KITTI dataset: KITTI stereo and KITTI VO, which both consist of challenging and varied road scene imagery collected from a test vehicle. Ground truth depth maps are obtained from 64-line LIDAR data. The main difference between these two dataset is that the ground truth depth maps for the stereo dataset were manually corrected and interpolated based on the neighboring frames. Note we only used the lower half part of the images ($200\times 1226$) as the upper half part generally include large part of sky and there is no depth measurements available. The input of our experiment was synthesized low-resolution LIDAR points which was downsampled by a factor of 6 in horizontal direction and 3 in vertical direction.

\begin{figure}[t]
\begin{center} 
 \subfigure[Unary energy versus the number of iterations.]{
\includegraphics[width=0.45\columnwidth]{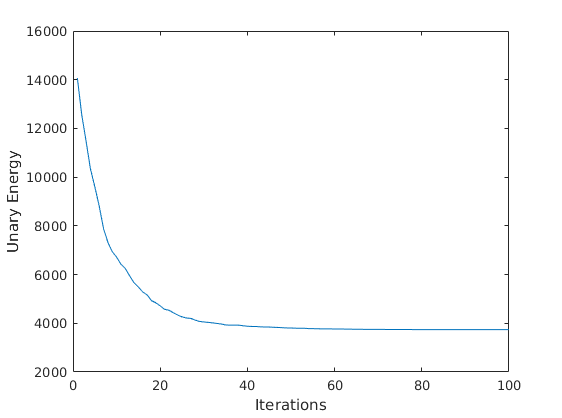} }
 \subfigure[Total energy versus the number of iterations.]{
   \includegraphics[width=0.45\columnwidth]{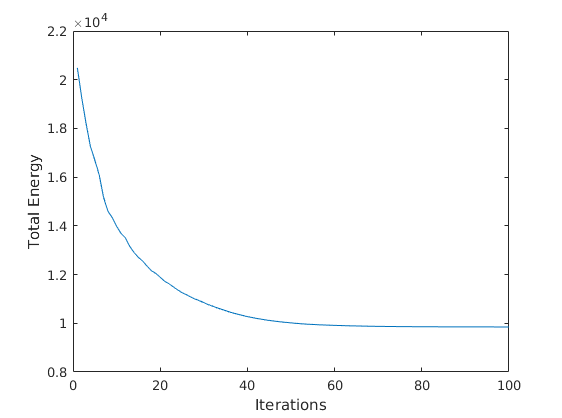} }
 \caption[Energy distribution.]{\label{fig:energyDist} This figure shows the energy with respect to the number of iterations in CRF based method.}
 \end{center}
\end{figure}

\noindent
\textbf{Piece-wise planar method setting:} As this method performs on superpixel level, we use SLIC \cite{SLIC2012} segmentation method to provide segments and the number of super-pixels is manually set to 800. The number of particles is the set to 10. The total iteration of PCBP is set to 40 as in Fig.~\ref{fig:energyDist} as the energies - both total and unary one - become flat since then.

\noindent
\textbf{``Cardboard world'' method setting:} We use SLIC \cite{SLIC2012} segmentation method to provide segments and the number of super-pixels is manually set to 1200. In the TRW-s inference step, the number of particles is set to 10, and divided into 3 parts: 1 particle represents the road plane, 4 particles are sampled from its neighbors and 5 particles are newly generated from MCMC process. The number of iteration of PCBP is set to 20 as there is only 1 parameter that need to optimize in this case.

\subsection{Error Metrics}
To evaluate the performance of depth interpolation, we use the following three quantitative metrics:
\begin{compactenum}[1)]
\item \textbf{Mean relative error} (MRE), which is defined as:
\begin{equation}
\mathrm{e}_{MRE} = \frac{1}{N}\sum_{i=1}^N\frac{|\mathrm{d}_i-\widehat{\mathrm{d}_i}|}{\mathrm{d}_i},
\end{equation}
where $\mathrm{d}_i$ and $\widehat{\mathrm{d}_i}$ are the ground truth depth and inferred depth respectively. A lower $\mathrm{MRE}$ indicates a better dense depth prediction performance.

\item \textbf{Bad pixel ratio} (BPR) measures the percentage of erroneous positions in total, where a depth prediction result is determined as erroneous if the absolute depth prediction error is beyond a given threshold $\mathrm{d}_{th}$. In our experiment, we set the bad pixel threshold as $\mathrm{d}_{th} = 3$ meters in VO dataset. A lower bad pixel ratio indicates a better depth prediction results. 

\item \textbf{Mean absolute error} (MAE) is defined as:
\begin{equation}
\mathrm{e}_{MAE} = \frac{1}{N}\sum_{i=1}^N|\mathrm{d}_i-\widehat{\mathrm{d}_i}|,
\end{equation}
where $d_i$ and $\widehat{\mathrm{d}_i}$ are the ground truth depth and depth prediction respectively. A lower mean absolute error indicates a better dense depth prediction performance achieved. It also indicates the average depth estimation error in meters.
\end{compactenum}
Bad pixel ratio, mean relative error and mean absolute error measure different statistics of the dense depth prediction results, which jointly evaluate the prediction performance.

\subsection{Experiment Results}
\begin{figure}[t]
\begin{center} 
 \subfigure{
\includegraphics[width=0.3\columnwidth]{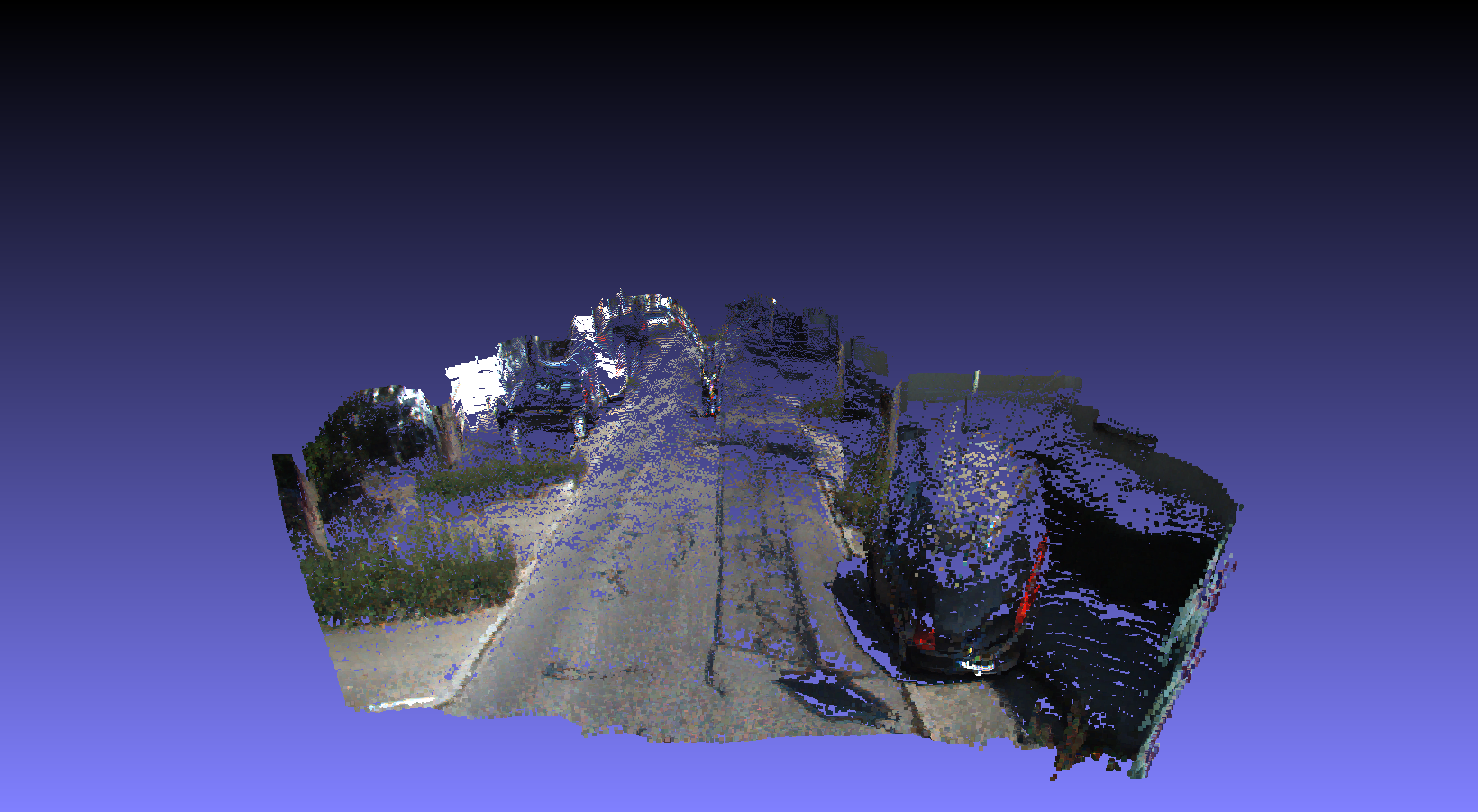} } \hspace{-.3cm}
\subfigure{
\includegraphics[width=0.3\columnwidth]{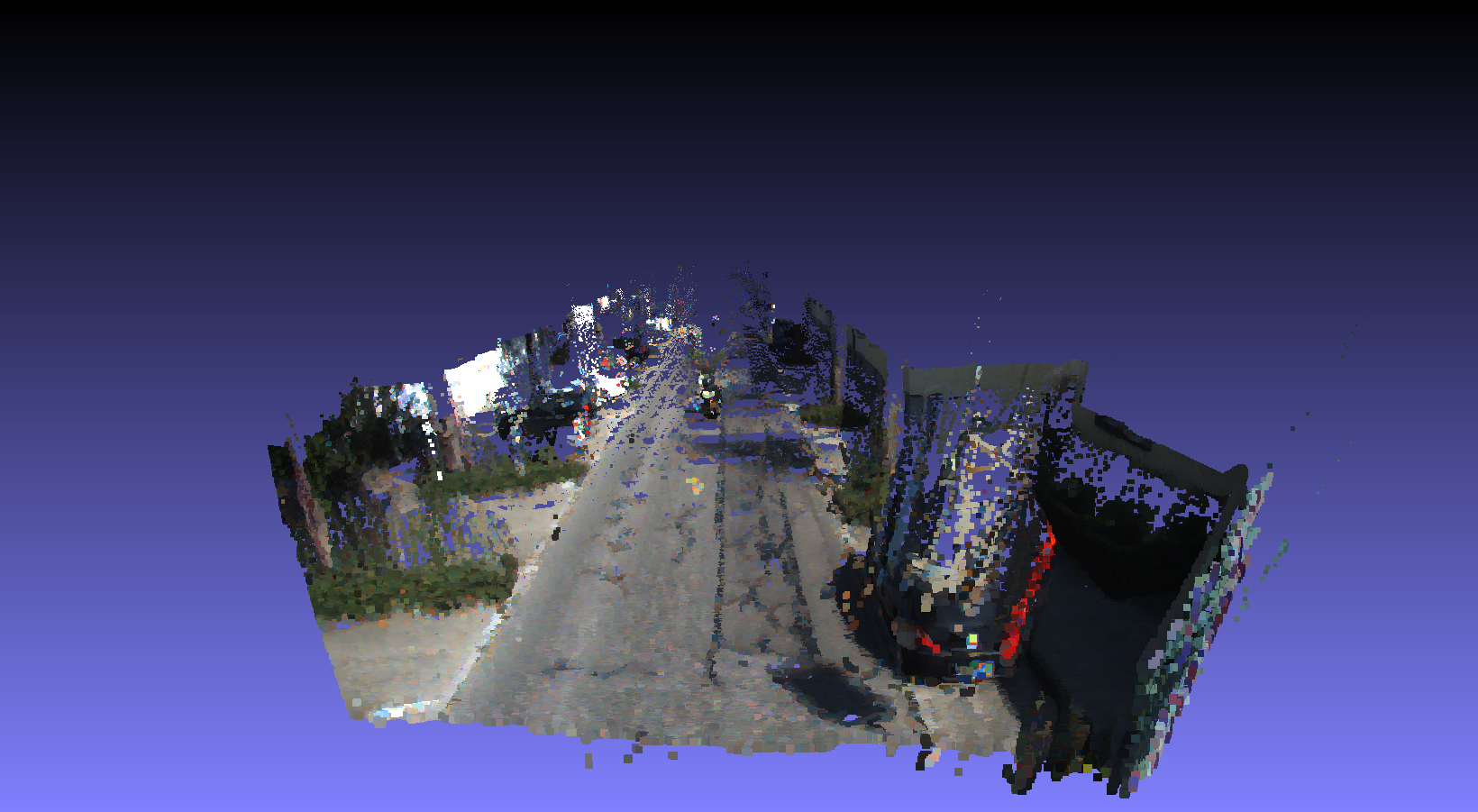} } \hspace{-.3cm}
\subfigure{
\includegraphics[width=0.3\columnwidth]{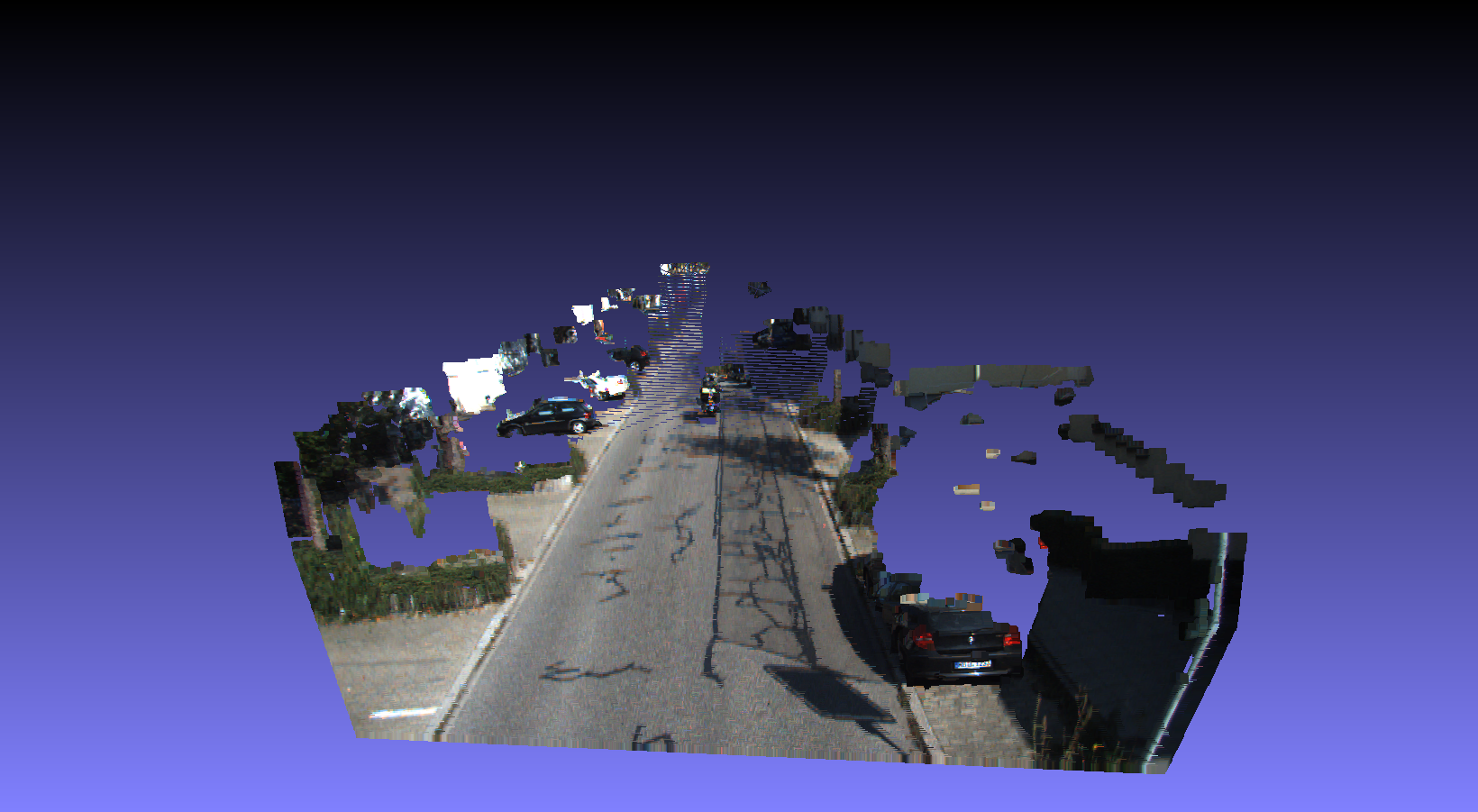} }\vspace{-.3cm}
\subfigure{
\includegraphics[width=0.3\columnwidth]{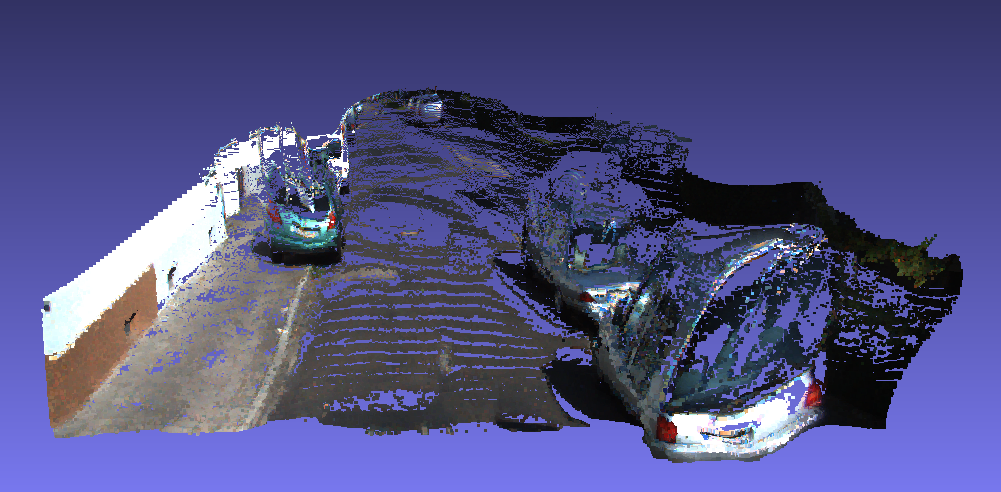} } \hspace{-.3cm}
\subfigure{
\includegraphics[width=0.3\columnwidth]{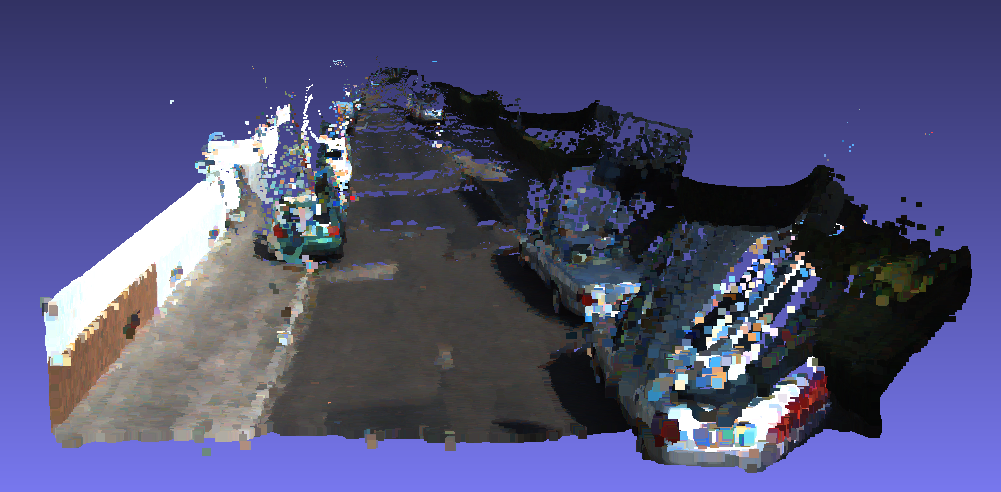} } \hspace{-.3cm}
\subfigure{
\includegraphics[width=0.3\columnwidth]{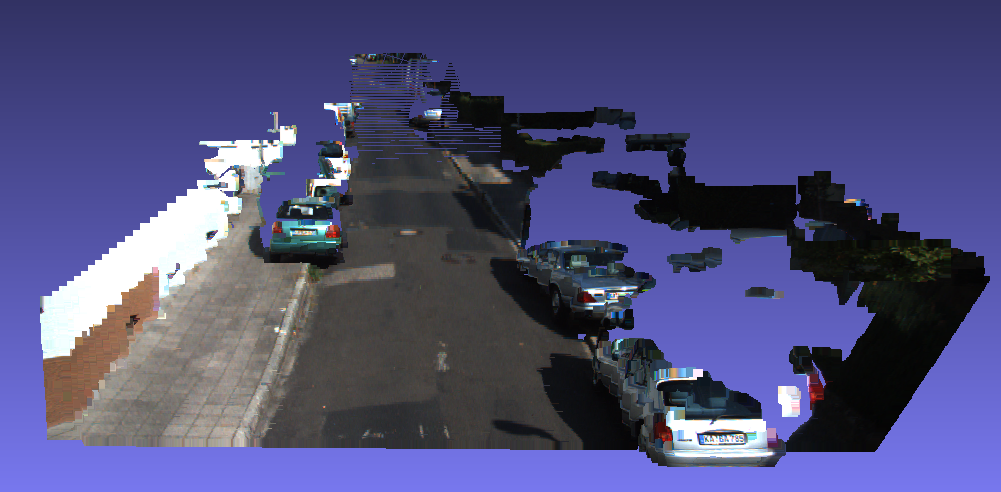} }\vspace{-.3cm}
\subfigure{
\includegraphics[width=0.3\columnwidth]{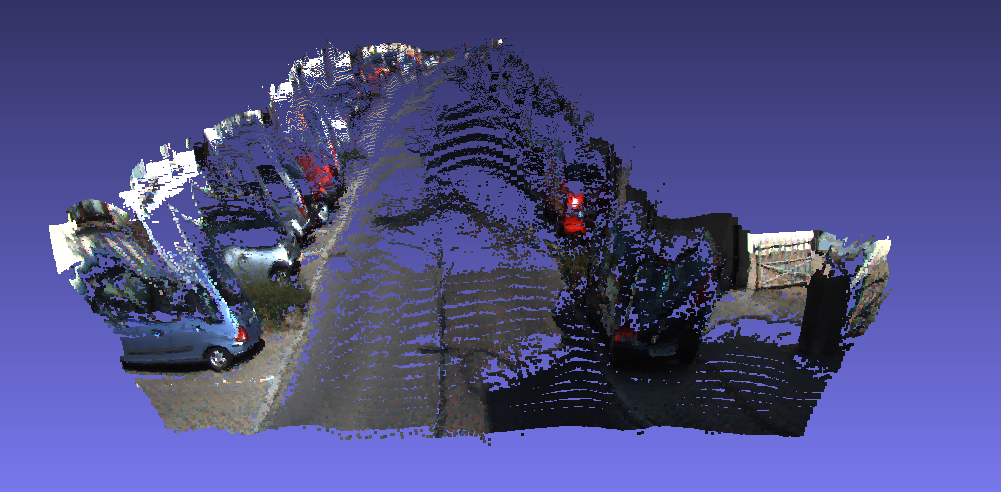} } \hspace{-.3cm}
\subfigure{
\includegraphics[width=0.3\columnwidth]{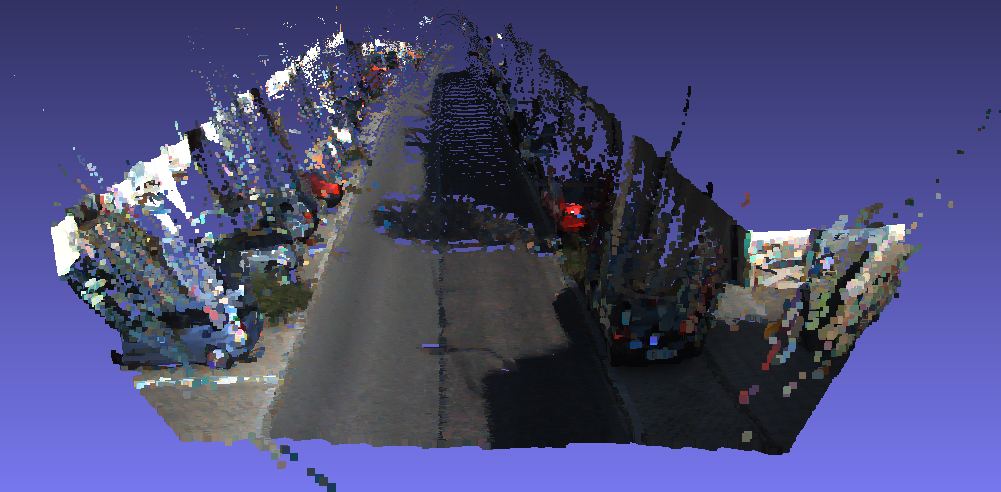} } \hspace{-.3cm}
\subfigure{
\includegraphics[width=0.3\columnwidth]{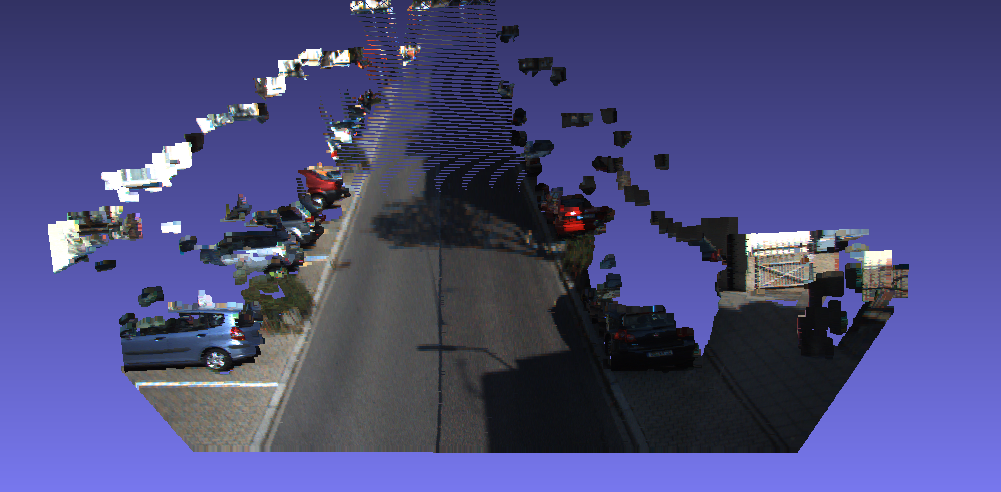} }\vspace{-.3cm}
\subfigure{
\includegraphics[width=0.3\columnwidth]{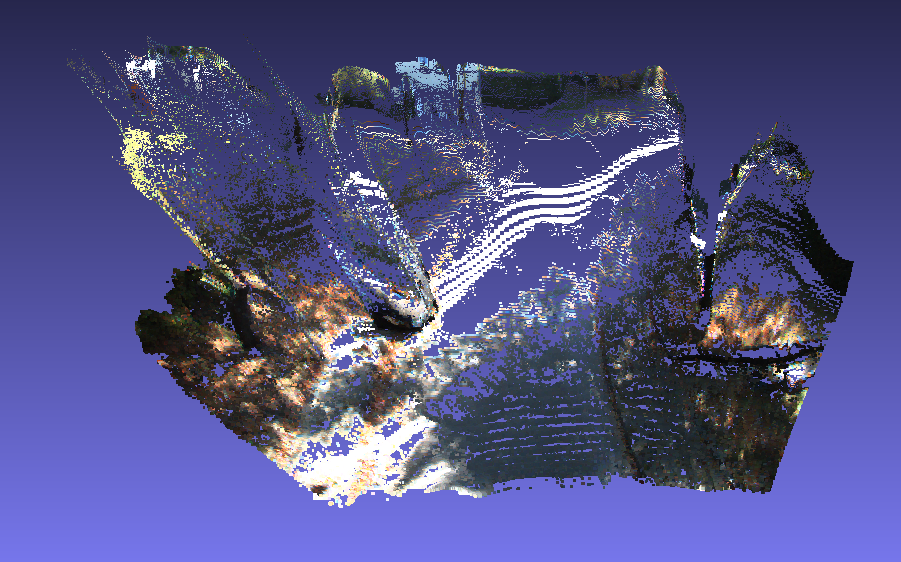} } \hspace{-.3cm}
\subfigure{
\includegraphics[width=0.3\columnwidth]{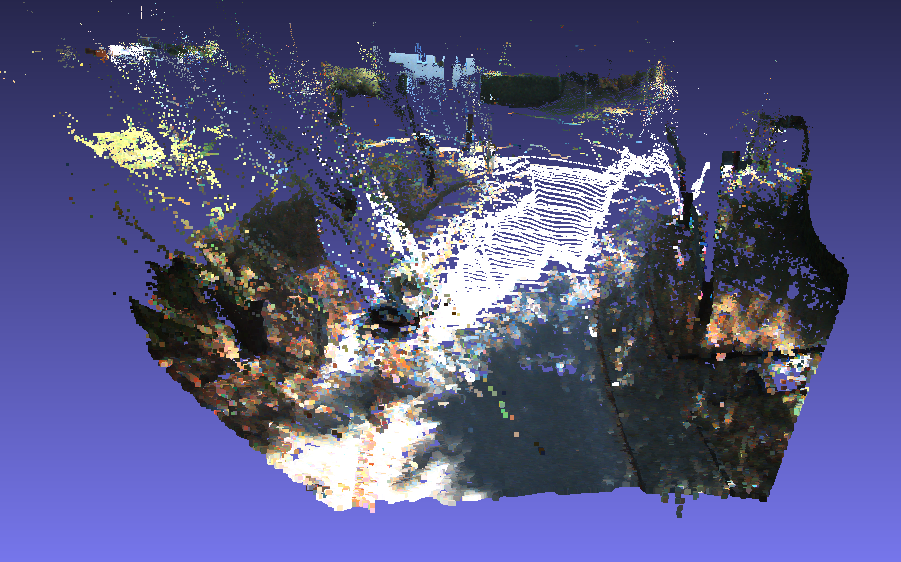} } \hspace{-.3cm}
\subfigure{
\includegraphics[width=0.3\columnwidth]{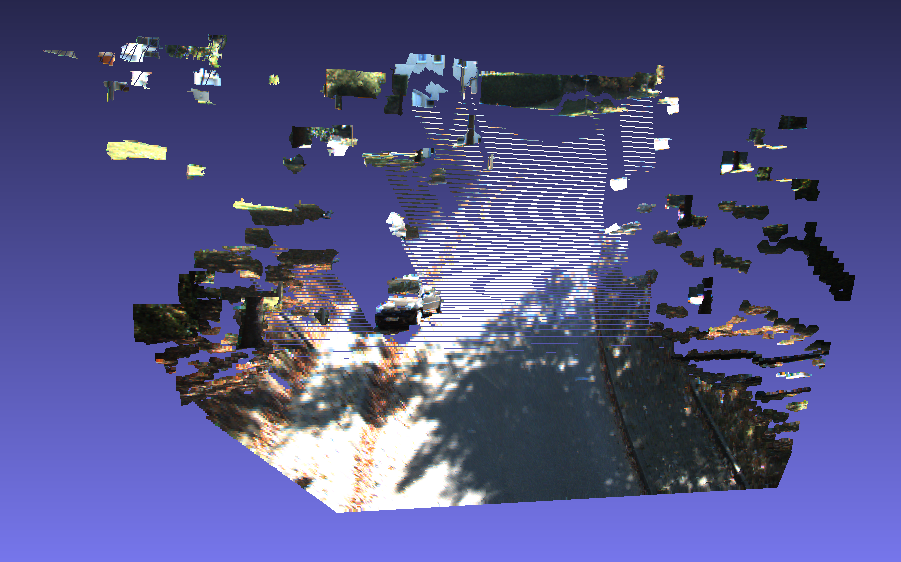} }
\caption[Comparison in coloured 3D point clouds.]{\label{fig:p3} \textbf{Comparison in coloured 3D point clouds.} Left column: results from \cite{Barron2016}; Middle column: results from PCA based colour guide method; Right column: results from the ``cardboard world'' method.}
 \end{center}
\end{figure}

Our quantitative results are shown in Table~\ref{kittivo}. The piece-wise planar method outperforms all the other methods. However, since our goal is to generate both useful and visual pleasant 3D point cloud, we also provide qualitative results as shown in Fig.~\ref{fig:p3}. For better comparison, we compare our method with several other state-of-the-art depth super-resolution methods: bilateral solver\cite{Barron2016}, as well as our colour-guided PCA based depth interpolation method. The initial depth map used for the bilateral solver \cite{Barron2016} was generated by a general smooth interpolation method \cite{Garcia20101167}.

\begin{table}[!htb]
\centering
\caption[Evaluation on the KITTI VO dataset.]{Evaluation on the KITTI VO dataset.}
\label{kittivo}
\tabcolsep=0.38cm
\begin{tabularx}{\columnwidth}{c|c|c|c|c}
\hline
        & Bilateral\cite{Barron2016}  & colour PCA        & Piece-wise      & ``Cardboard"\\ \hline
MRE(\%)     & 7.36                        & 5.73             & \textbf{4.87}   & 7.85 \\ \hline
BPR(\%)     & 9.46                        & 7.51             & \textbf{5.82}   & 8.57 \\ \hline
MAE(m)	& 1.20                        & 1.04             & \textbf{0.80}   & 1.29 \\ \hline
\end{tabularx}
\end{table}

As we can observe, bilateral solver \cite{Barron2016} provides over-smoothed results, large distortion can be observed in the areas with different colours. For example, in Fig.~\ref{fig:p3}, when there are shadows on road, it tends to assign same depth to same colour areas when lacking information, therefore create stripes effects in 3D point clouds. However, in our PCA based colour guided method, with the help of high order smoothness term and the PCA bases as a global constrain, it shows some resistances to the misleading of false boundaries that introduced by the colour images. Also, it can recover the shape of cars. However, as it is a pixel-wise algorithm, it is hard to estimate every pixel with the right depth. Therefore, we can find there are holes on the road plane. This can be harmful for autonomous driving system as it creates many false road pits and/or false obstacles.

\begin{figure}[!htp]
\begin{center} 
 \subfigure{
\includegraphics[width=0.48\columnwidth]{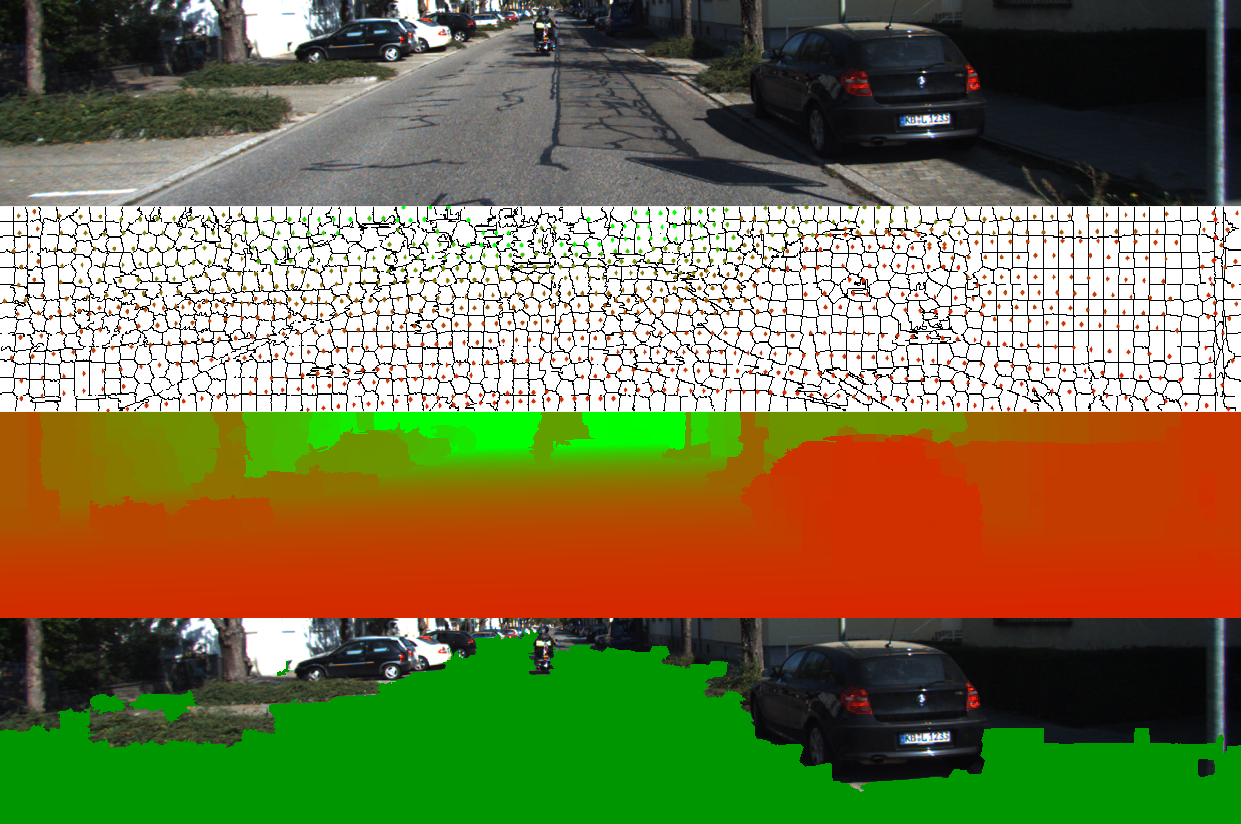} } \hspace{-.3cm}
\subfigure{
\includegraphics[width=0.48\columnwidth]{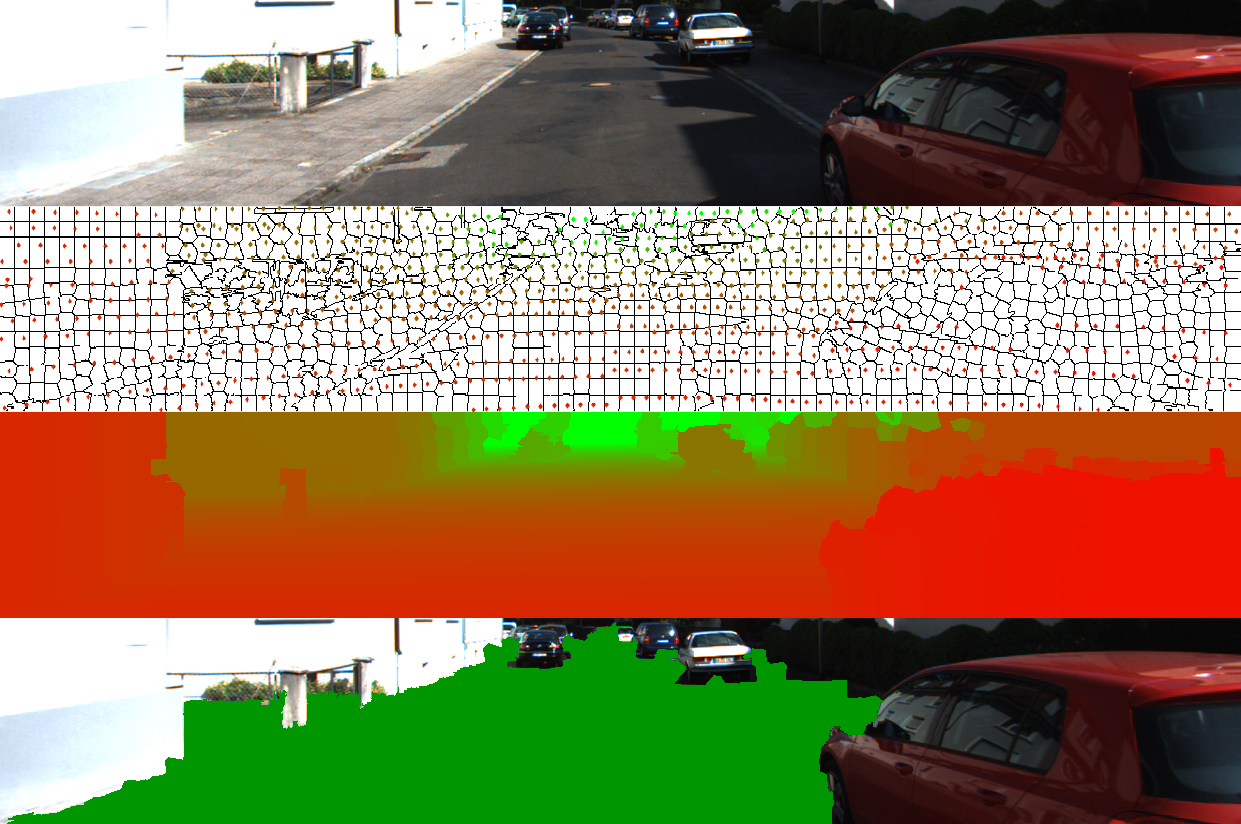} }
\caption[Results of ``Cardboard world'' method.]{\label{fig:pcVO1}\textbf{Example results of ``Cardboard world'' method:} Top to bottom: reference colour frames, inputs of our method, recovered depth map and segmented free space (coloured with green).}
 \end{center}
 \vspace{-4mm}
\end{figure}

The ``cardboard world'' algorithm, on the other hand, does not have these drawbacks. As we can see from all these results, none of the road plane was fooled by shadows or marks. For better illustration of our algorithm, in Fig.~\ref{fig:pcVO1}, we provide the input and output of our algorithm and the road plane segmentation as well.  From top to bottom, each figure consists of the reference colour image, the input of our algorithm (LIDAR measurements and super-pixel segmentation), our recovered dense depth map and our road plane segmentation result correspondingly. \emph{Note that the colour image is only used to generate the super-pixel segmentation and only sparse LIDAR measurements are used in the optimization.}

As we can see from the figures, all dominant road plane space (labeled by dark green) have been accurately extracted. In typical city road scenarios (i.e., Fig.~\ref{fig:pcVO1}), our method can successfully extract cars that are parked on side road and a motorbike over 22 meters away from only 3 LIDAR points on it.

\begin{figure}[!htp]
\begin{center} 
\subfigure[\scriptsize MRE:4.9 BPR:6.1 MAE:0.846 ]{
\includegraphics[width=0.48\columnwidth]{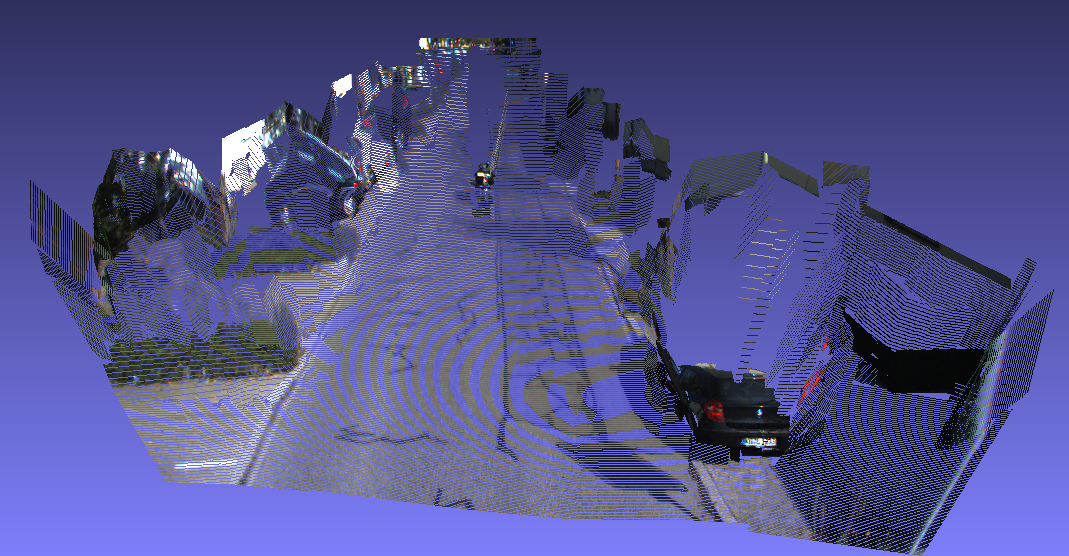} } \hspace{-.3cm}
 \subfigure[\scriptsize MRE:6.5 BPR:8.7 MAE:1.041]{
\includegraphics[width=0.48\columnwidth]{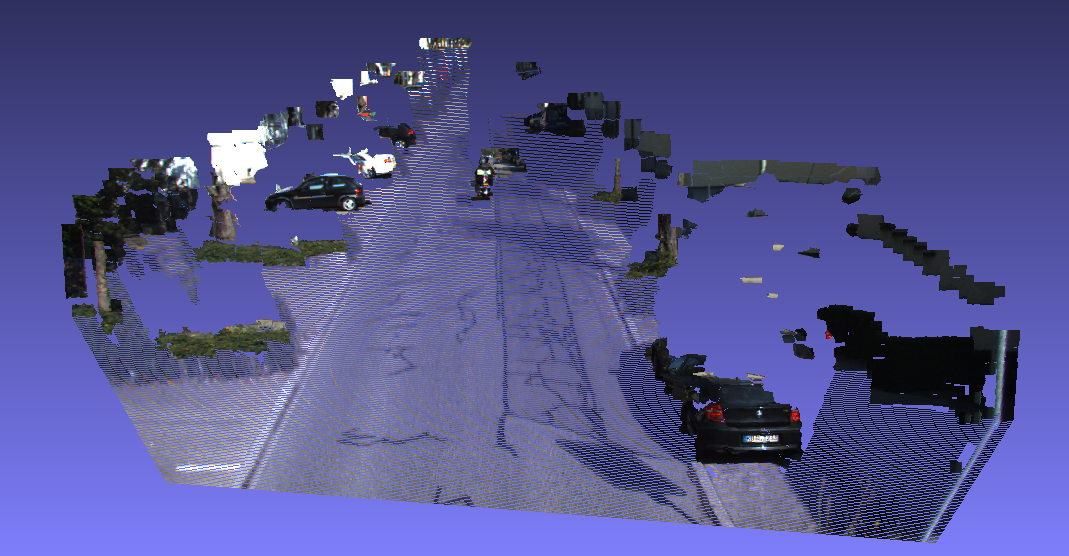} }\vspace{-.3cm}
\subfigure[\scriptsize MRE:5.2 BPR:4.2 MAE:0.583]{
\includegraphics[width=0.48\columnwidth]{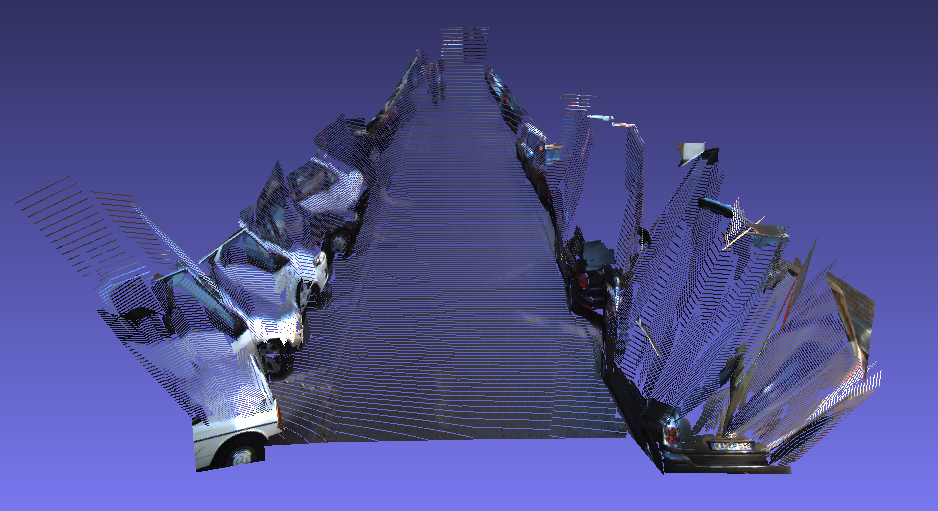} } \hspace{-.3cm}
 \subfigure[\scriptsize MRE:9.5 BPR:6.2 MAE:1.060]{
\includegraphics[width=0.48\columnwidth]{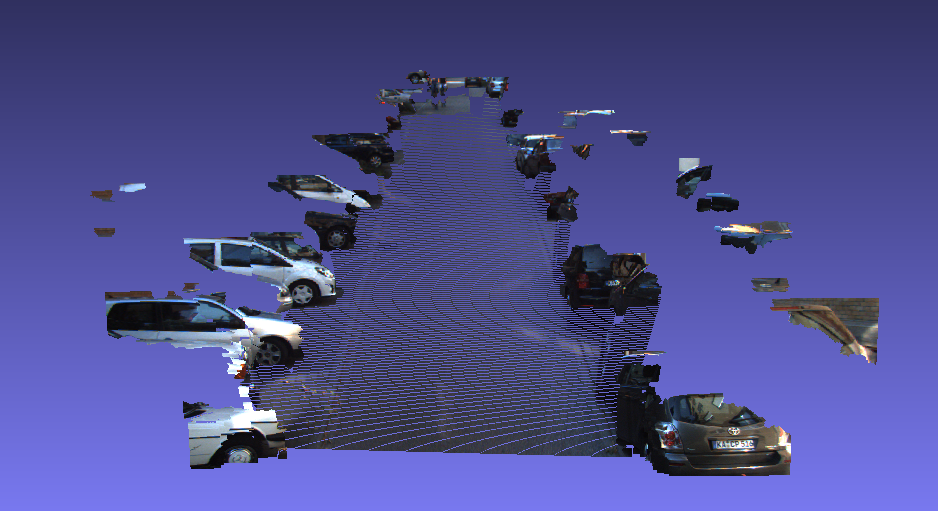} }
 \caption[CRF based method \vs ``Cardboard world'' method.]{\label{fig:cp1} Comparison between CRF based method (left) and ``cardboard world'' method (right).}
 \end{center}
  \vspace{-4mm}
\end{figure}

To better illustrate the advantage of our ``cardboard world'' model over the standard piece-wise planar model, we also provided quality comparison between them. As we can see from Fig.~\ref{fig:cp1}, both methods successfully recover road plane with resistant to shadows. The quantitative results show that the CRF based method achieves better performance. However, in turns of the quality of 3D point cloud, the CRF method largely distorts the shape of cars as in (a) and (g) while the ``cardboard world'' method provides less distorted, much clean and visually look better results. 

\subsection{Processing time}
Unlike pixel-level colour guided methods where the processing time largely depends on the input image resolution, the processing time of our method depends on the number of superpixels, the number of particles and the number of parameters to optimize. The higher the numbers, the longer the processing time is. In piece-wise planar model based solution, when we use 800 superpixels, 10 particles for each superpixel, the running time is about 10 seconds on average. However, in the more constrained case, our ``cardboard world'' method with more superpixels only needs 1 seconds on average. 

\section{Conclusion}
This paper aims at tackling the challenging task of predicting dense depth maps from very sparse measurements, we have proposed two different methods by exploits various local and global constraints inside the problem. The first method is based on the piecewise planar model of the scene, where dense depth prediction is reformulated as the optimization of planar parameters. The second method enforces strong regularization on the scene model to exploit the structural information in outdoor traffic scenes, which is more suitable for autonomous driving tasks. Unlike existing depth super-resolution methods that can be easily misled by marks or shadows on road, our methods inherently resist to these false guidance. Experimental results on the KITTI VO dataset show that our methods can efficiently recover dense depth map from less than $1000$ LIDAR points without losing important information for autonomous driving, i.e., obstacles on road, or creating false obstacles that may misleading self-driving vehicles. In future, we plan to exploit the temporal information in constraining the dense depth maps.

\bibliographystyle{unsrt}
\bibliography{egbib}

\end{document}